# 3D-DETNet: a Single Stage Video-Based Vehicle Detector

Suichan Li

School of Information Science and Technology, University of Science and Technology of China,
Hefei 230026, China.

## ABSTRACT

Video-based vehicle detection has received considerable attention over the last ten years and there are many deep learning based detection methods which can be applied to it. However, these methods are devised for still images and applying them for video vehicle detection directly always obtains poor performance. In this work, we propose a new single-stage video-based vehicle detector integrated with 3DCovNet and focal loss, called 3D-DETNet. Draw support from 3D Convolution network and focal loss, our method has ability to capture motion information and is more suitable to detect vehicle in video than other single-stage methods devised for static images. The multiple video frames are initially fed to 3D-DETNet to generate multiple spatial feature maps, then sub-model 3DConvNet takes spatial feature maps as input to capture temporal information which is fed to final fully convolution model for predicting locations of vehicles in video frames. We evaluate our method on UA-DETAC vehicle detection dataset and our 3D-DETNet yields best performance and keeps a higher detection speed of 26 fps compared with other competing methods.

## 1. INTRODUCTION

Video-based vehicle detection for driver assistance and traffic surveillance has received considerable attention over the last ten years. Recently, deep learning based approaches which can be roughly divided into two streams including single-stage detectors[1,2,3] and two-stage detectors[4,5,6], have achieved excellent results in still images. Compared to two-stage methods, single-stage methods always yield faster performance in speed which is more suitable for the real-time application. However, it is challenging to apply these methods for video object detection directly. The deteriorated object appearance invades which are seldom observed in still images, such as motion blur, video defocus, can damage recognition accuracy[7].

Nevertheless, the video has rich information about the same object instance. Such temporal information has been exploited in existing video classification and recognition methods [8,9]. These methods learn spatiotemporal features using 3D ConvNets. Du Tran et al.[9] found that 3D ConvNets were more suitable for spatiotemporal feature learning compared to 2D ConvNets. On the other hand, recent work on single-stage detectors, such as YOLO[1,2] and SSD[3] yield faster performance in speed, but with lower accuracy relative to state-of-the-art two-stage methods. In [10], the central cause of why single-stage approaches have trailed the accuracy of two-stage detectors thus far was investigated, and a focal loss was proposed to address the problem.

As motivated by the success of 3D ConvNets[8,9] and focal loss[10], we propose a new single-stage video-based vehicle detector integrated with 3DCovNet and focal loss, called 3D-DETNet, to improve the per-frame vehicle detection performance and keep high detection speed at the same time. Our 3D-DETNet can be trained in an end to end fashion, and does not need training in multistage paradigm as [4,5] follow. Our approach is evaluated on the recent DETRAC vehicle detection dataset[11]. Rigorous evaluating study has verified that our method is effective and significantly improves upon single frame baselines. In addition, we compared it with other detectors and the results showed that our approach has better performance both in accuracy and inference time.

## 2. RELATED WORKS

### 2.1 Hand-engineered feature based detectors

Early works of vehicle detection used Haar-like features and Histograms of Oriented Gradient (HOG)[12] to detect vehicles in images. In [13], along with the histogram of oriented gradients (HOG), the authors proposed and implemented a new type of feature vector, i.e., HOG symmetry vectors for vehicle detection. The feature pyramids were also used to detect objects in [14]. In [15], a mixture of multi-scale deformable part models was used to represent highly variable objects. Nevertheless, the success of those approaches generally depends on the stability of data representation features, like moving objects' scale changes, translation, etc.[16]

## 2.2 Deep learning based object detectors

With the development of modern deep ConvNets, object detectors like R-CNN[4] and YOLO[2] showed dramatic improvements in accuracy. Based on a region proposal strategy, R-CNN[4] applies high capacity convolution neural networks (CNNs) to bottom-up region proposals to locate objects in image. YOLO[2] generates several candidate objects for anchor boxes in the image and locates the objects in a single shot. These methods based on convolution neural networks can automatically extract robust features without human assistance and achieve excellent results in still images. However, directly applying them for video object detection is challenging, due to motion blur, video defocus, etc.

## 2.3 Learning temporal features with 3D ConvNets

Recently, 3DConvNets were proposed to learn spatial-temporal features for video classification and recognition methods. In [8], a 3D CNN model was developed for human action recognition and 3D ConvNets were proposed to extract features both in spatial and temporal dimensions. Du Tran et al.[9] found that 3D ConvNets were more suitable for spatiotemporal feature learning compared to 2D ConvNets. More recently, following the philosophy of ResNet[17], Qiu, Z et al.[18] proposed a new architecture, named Pseudo-3D Residual Net and achieved clear improvements on multiple benchmarks.

## 3. MODEL ARCHITECTURE

We illustrate our 3D-DETNet in Figure 1 and as follows:

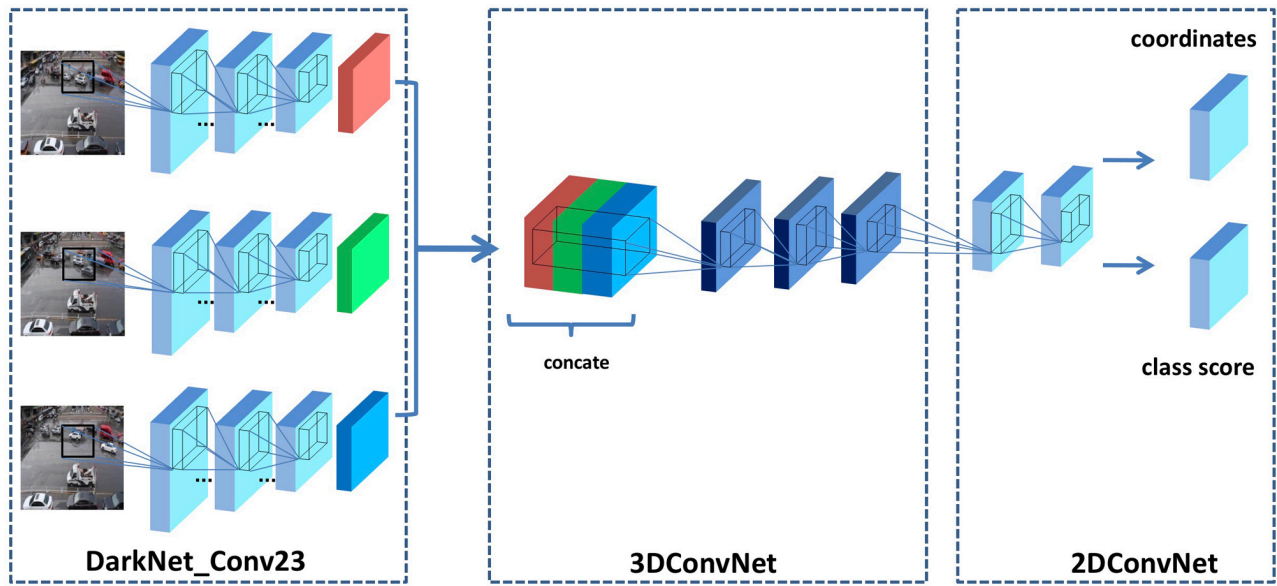

Figure 1.The architecture of our 3D-DETNet. Three sub-models are involved including DarkNet_Conv23, 3D Convolution Networks (3DConvNet) and 2D Convolution Networks (2DConvNet). The DarkNet_Conv23 is responsible for generating rich spatial feature representations from multiple frames. The 3DConvNet is mainly to capture the motion or temporal information encoded in multiple contiguous video frames. The 2DConvNet takes the feature maps from the previous 3D convolution Network (3DConvNet) as input and predicts final outputs.

**DarkNet-Conv23**：As shown in Figure 1, the multiple contiguous frames are initially fed to DarkNet-Conv23 [2] to generate rich spatial features representations respectively. The DarkNet-Conv23 is pre-trained in ImageNet dataset[19] and is capable of extracting rich spatial features. The output feature maps are sent to 3DConvNet model afterwards for further extraction of temporal features.

**3DConvNet:** The sub-model 3DConvNet is mainly to capture the motion or temporal information encoded in multiple contiguous video frames. However different from[8, 9], our 3DConvNet model takes the spatial feature maps from the previous DarkNet-23 model as input, instead of taking multiple video frames as input directly.

**2DConvNet:** The 2DConvNet which consists of several 2D convolution layers takes the feature maps from the previous 3D convolution Network (3DConvNet) as input. The model's final outputs include bounding box coordinates and class probability.

We describe below each individual model in details.

### 3.1 Darknet-Conv23 for spatial feature

Recently, VGG-16 [20] was leveraged by most of detection method as the feature extractor base. VGG-16 is powerful but complex. In [2], a new image classification network model called Darknet-19 was proposed, which is faster and requires fewer parameters than VGG-16. Darknet-19 has 19 convolution layers and 5 maxpooling layers, it was trained on IamgeNet dataset[19], and achieved 72.9% top-1 accuracy as well as 93.3% top-5 accuracy. We leverage DarkNet-19 as our feature extractor base by removing the last convolution layer, which called DarkNet_Conv23 (the final model has 23 layers), so our detector can benefit from extra big dataset, which is a common practice for transfer learning.

### 3.2 3DConvNet for temporal feature

As mentioned earlier, video vehicle detection has a greater challenge than vehicle detection in static images. We want to extract motion information to overcome the challenge of video blur and so on. To this end, we propose to perform 3D convolutions in the convolution stages of CNNs to compute features in temporal dimensions and get the motion information. Similar to the case of 2D convolution, we can apply multiple 3D convolutions with distinct kernels to the same location in the previous layer to extract multiple types of features.

Formally, the value at position (x, y, z) on the jth feature map in the ith layer can be given by

$$v_{i,j}^{x,y,z} = f(\sum_m \sum_{p=0}^{P_i} \sum_{q=0}^{Q_i} \sum_{r=0}^{R_i} w_{i,j,m}^{p,q,r} v_{i-1,m}^{x+p,y+q,z+r} + b_{i,j}) \quad (1)$$

where $Q_i$, $R_i$ are the size of the 3D kernel along spatial dimension respectively, $P_i$ is the size of the 3D kernel along temporal dimension, $w_{i,j,m}^{p,q,r}$ is the (p, q, r)th value of the kernel connected to the mth feature map in the previous layer. $b_{i,j}$ is bias term.

Notably, our 3DConvNet model does not take multiple video frames as input directly, but instead take the spatial feature maps from the previous DarkNet-23 model as input. To some extent, our 3DConvNet plays the role of feature fusion or aggregation.

Based on the 3D convolution described above, our 3DConvNet architecture can be devised. As shown in Figure 1, our 3DConvNet is composed of three 3D convolution layers. According to the findings in [9], small receptive fields of 3x3x3 convolution kernel yield best result. Hence, for our architecture, we set the kernel size to 3x3x3, 1x1x1, 3x3x3 with appropriate strides and padding size respectively.

### 3.3 2DConvolution prediction for detection

We use the convolution layer predicting both class probabilities and bounding box coordinates, instead of fully connected layer. Our 2DConvNet model consists of two 3x3 convolution layer with 1024 filters each followed by a 1 x 1 convolution layer, which has the number of outputs we need for vehicle detection. Follow YOLO [2], we use anchor boxes to predict bounding boxes. We don't hand pick the bounding box dimensions, i.e. width and height, instead we run k-means clustering on the DATRAC dataset to find good priors. Finally, we choose k=5 and get 5 anchor box priors.

Our model predicts 5 bounding boxes at each cell in the output feature map, and for each bounding box we predict 5 coordinates: *x*, *y*, *w*, *h* and confidence *c*. Like [2], we don't regress the coordinates relative to original frames, instead we parameterize the coordinates $(b_x, b_y, b_w, b_h, b_c)$ by $(t_x, t_y, t_w, t_h, t_c)$:

$$\begin{aligned} b_x &= \sigma(t_x) + c_x \\ b_y &= \sigma(t_y) + c_y \\ b_w &= p_w e^{t_w} \\ b_h &= p_h e^{t_h} \\ b_c &= \sigma(t_o) \end{aligned} \quad (2)$$

where $(c_x, c_y)$ is offset of cell from the top left corner of the image, $(p_w, p_h)$ are width and height of anchor priors respectively, $\sigma(\cdot)$ is a logistic activation function in range [0, 1]. $(b_x, b_y, b_w, b_h, b_c)$ is final prediction coordinates.

### 3.4 Focal loss and multi-part loss function

The state-of-the-art object detectors are based on a two-stage approach popularized by R-CNN[3]. One-stage detectors yield faster but with accuracy within 10-40% relative to state-of-the-art two-stage methods[10]. The central cause of why one-stage approaches have trailed the accuracy of two-stage detectors thus far was investigated, and a focal loss was proposed to address the problem in [10]. Formally, the focal loss can be defined as:

$$FL(p_t) = -\alpha_t (1 - p_t)^\gamma \log(p_t) \quad (3)$$

where $p_t$ which belongs to [0, 1] is the model's estimated probability for the class with positive label. $\gamma \geq 0$ is a tunable focusing parameter. See Figure 2, the focal loss is visualized for multiple values of $\gamma$.

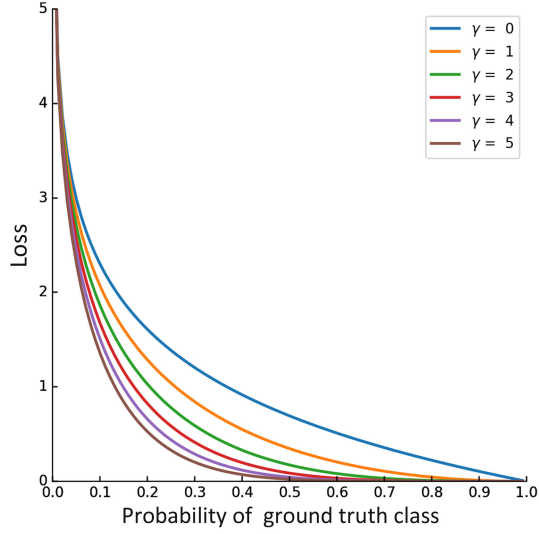

Figure 2. Visualization of focal loss for multiple values of $\gamma$.

As motivated by focal loss, we leverage the focal loss for our classification loss. As mentioned earlier, our predictive output includes coordinates and categories information, so our loss function is a multi-part loss and can be formalized as following

$$L = \lambda L_{loc}(t, s) + L_{cls}(s) \quad (4)$$

where $L_{loc}(t, s)$ and $L_{cls}(s)$ are the location regression loss and classification loss. $\lambda$ is a balance factor. For regression loss, we use smooth L1 loss[5] defined as

$$smooth_{L1} = \begin{cases} x^2 & \text{if } |x| < 1 \\ |x| - 0.5 & \text{otherwise} \end{cases} \quad (5)$$

For classification loss, we use focal loss as defined in Equation (3).

### 3.5 Network Training

Follow YOLO[2], we only want one bounding box predictor to be responsible for each object. So we assign one predictor as positive bounding box which has the highest IOU with the ground truth.

3D-DETNet is trained with stochastic gradient descent (SGD) for about 80 epochs on training sets. We set learning rate to 10e-3 for first 60 epochs and then decrease to 10e-4 for remaining epochs. For data augmentation, we use random crop and horizontal image flipping. We also randomly adjust the exposure and saturation of the image in HSV color space. We use the mini-batch size of 32, a weight decay of 0.0005 and momentum of 0.9.

Due to limitation of memory, we choose three frames as input of 3D-DETNet during training. Note that the neighbor frames are randomly sampled from a range [-10, 10] relative to the reference frame.

## 4. EXPERIMENTS

We evaluate the experiments on the UA-DETRAC vehicle detection dataset[11]. The dataset consists of 10 hours of videos captured in different scenarios including sunny, cloudy, rainy and night. There are more than 140 thousand frames in the UA-DETRAC dataset and 8250 vehicles that are manually annotated, leading to a total of 1.2M labeled bounding boxes of objects.

To evaluate the effectiveness of our detect framework, we conducted three experiments in this paper. The first experiment is evaluated on a small validation dataset to compare focal loss with cross-entropy loss and to find a best $\gamma$ for focal loss. The second one evaluate the performance of our 3D-DETNet on validation dataset. The final one is evaluated in the DETRAC test dataset and compared with other competing methods.

### 4.1 Focal loss vs. cross-entropy loss

In this section, we compare focal loss with cross-entropy loss which is utilized in many classification problems. On the other hand, focal loss has a tunable parameter $\gamma$, so we evaluate focal loss with multiple values of $\gamma$ and find a best gamma for our task. The results are summarized in Table 1.

Table 1. The mean Average Precision (mAP) of cross-entropy and focal loss with different $\gamma$ on small validation dataset

| Loss | $\gamma$ | mAP |
|---|---|---|
| Cross-entropy | 0 | 81.52 |
| Focal loss | 1 | 82.16 |
| Focal loss | 2 | **83.85** |
| Focal loss | 3 | 83.64 |
| Focal loss | 4 | 83.29 |

From the result, it can be seen that the focal loss yields better performance compared to standard cross-entropy loss which accords with the study in [10]. For focal loss, we observe that $\gamma = 2$ gets the best result for our task. For remaining experiments, we set $\gamma = 2$ for focal loss.

### 4.2 Evaluate the effectiveness of 3D-DETNet

This experiment is mainly to evaluate the effectiveness of our 3D-DETNet. The 3DConvNet is proposed to capture the motion information encoded in multiple contiguous frames in our framework. If we remove the 3DConvNet in our framework, the remaining network architecture is similar to YOLO[2]. To evaluate of effectiveness our methods, we compare 3D-DETNet with YOLO[2] on validation dataset. The Table 2 show the performance of our 3D-DETNet and YOLO on validation set in different scenarios.

Table 2. The overall mAP in validation dataset.

| Method | Overall | Sunny | Cloudy | Rainy | Night |
|---|---|---|---|---|---|
| YOLO | 80.23 | 83.28 | 84.19 | 67.98 | 77.35 |
| 3D-DETNet | **82.86** | **84.32** | **85.50** | **70.26** | **80.23** |

Quantitatively speaking, our 3D-DETNet performs better than YOLO. We note that our 3D-DETNet achieves overall mAP 82.86, as comparison, the YOLO has a lower overall mAP of 80.23. On the other hand, we note that our method outperforms YOLO by a larger margin under scenarios of rainy and night compared to sunny and cloudy scenarios, while rainy and night scenarios are more challenging than sunny and cloudy scenarios. These results indicate that our 3DConvNet has ability to capture strong temporal information and improve the effectiveness of per-frame vehicle detection.

### 4.3 Compare with other methods

Tabel 3. The mean Average Precision (mAP) and speed on DEATRAC test dataset generated by our method and other competing vehicle detection methods.

| Method | Overall | Easy | Medium | Hard | Cloudy | Night | Rainy | Sunny | Speed(fps) |
|---|---|---|---|---|---|---|---|---|---|
| DPM[15] | 25.70 | 34.42 | 30.29 | 17.62 | 24.78 | 30.91 | 25.55 | 31.77 | 0.17 |
| ACF[14] | 46.35 | 54.27 | 51.52 | 38.07 | 58.30 | 35.29 | 37.09 | 66.58 | 0.67 |
| RCNN[4] | 48.95 | 59.31 | 54.06 | 39.47 | 59.73 | 39.32 | 39.06 | 67.52 | 0.10 |
| Ours | **53.30** | **66.66** | **59.26** | **43.22** | **63.30** | **52.90** | **44.27** | **71.26** | **26** |

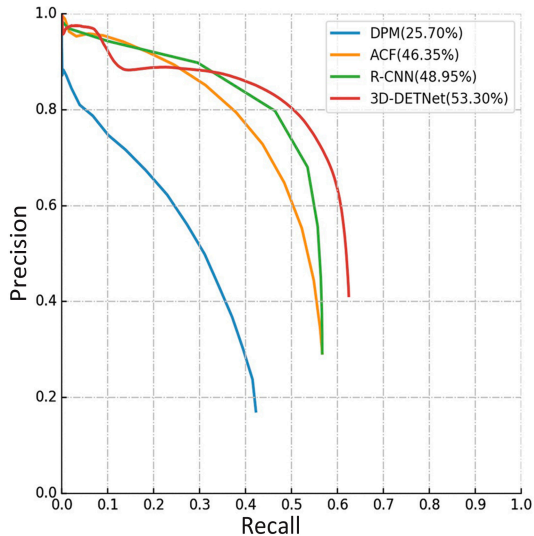

Figure 3. Precision-recall curve of different methods on test set.

In this experiment, we evaluate our model on the DETRAC official test dataset. We compare our method with three state-of-the-art vehicle detection approaches. The results are showed in table 3. Notably, our method performs the best on all subcategories and keeps a higher speed of 26 fps. We also report precision-recall curves in Figure 3. As shown in Figure 3, we note that our approach has better detection coverage as well as accuracy.

## 5. CONCLUSION

In this paper, we propose a new one-stage vehicle detector integrated with 3DCovNet and focal loss, called 3D-DETNet. Draw support from 3D Convolution and focal loss, our method has ability to capture motion information and is able to improve the per-frame vehicle detection performance. Compared with other competing methods, our 3D-DETNet yields best performance and run at 26 fps on a moderate commercial GPU. In the future work, we plan to further investigate the ability of 3D Convolution network and improve the performance of vehicle detection in video.

## REFERENCES

[1] Redmon, J., Divvala, S., Girshick, R. and Farhadi, A., "You only look once: Unified, real-time object detection," Proc. CVPR, 779-788 (2016).
[2] Redmon, J. and Farhadi, A., "YOLO9000: better, faster, stronger," arXiv preprint arXiv:1612.08242 (2016).
[3] Liu, W., Anguelov, D., Erhan, D., Szegedy, C., Reed, S., Fu, C. Y., et al., "SSD: Single shot multibox detector," Proc. ECCV 9905, 21-37 (2016).
[4] Girshick, R., Donahue, J., Darrell, T. and Malik, J., "Rich feature hierarchies for accurate object detection and semantic segmentation," Proc. CVPR, 580-587 (2014).
[5] Girshick, R., "Fast R-CNN," Proc. ICCV, 1440-1448 (2015).
[6] Ren, S., He, K., Girshick, R. and Sun, J, "Faster R-CNN: Towards real-time object detection with region proposal networks," Adv Neural Inf Process Syst., 91-99(2015).
[7] Zhu, X., Wang, Y., Dai, J., Yuan, L. and Wei, Y., "Flow-Guided Feature Aggregation for Video Object Detection," arXiv preprint arXiv:1703.10025(2017).
[8] Ji, S., Yang, M. and Yu, K., "3D convolutional neural networks for human action recognition," IEEE Trans. Pattern Anal. Mach. Intell., 35(1), 221-231(2013)
[9] Du, T., Bourdev, L., Fergus, R., Torresani, L. and Paluri, M., "Learning spatiotemporal features with 3d convolutional networks," Proc. ICCV, 4489-4497(2015).
[10] Lin, T. Y., Goyal, P., Girshick, R., He, K. and Dollár, P., "Focal loss for dense object detection," arXiv preprint arXiv:1708.02002(2017).
[11] Wen, L., Du, D., Cai, Z., Lei, Z., Chang, M. C., Qi, H., et al., "UA-DETRAC: A new benchmark and protocol for multi-object detection and tracking," arXiv preprint arXiv: 1511.04136 (2015).
[12] Dalal, N., Triggs, B., "Histograms of oriented gradients for human detection," Proc. CVPR 1,886-893(2005).
[13] Cheon, M., Lee, W., Yoon, C., Park, M., "Vision-based vehicle detection system with consideration of the detecting location," IEEE Trans Intell Transp Syst 13(3), 1243-1252 (2012).
[14] Dollár, P., Appel, R., Belongie, S., Perona, P, "Fast feature pyramids for object detection," IEEE Trans Pattern Anal. Machine Intell. 36(8), 1532-1545 (2014).
[15] Felzenszwalb P F, Girshick R B, McAllester D, et al., "Object detection with discriminatively trained part-based models," IEEE Trans Pattern Anal. Machine Intell. 32(9), 1627-1645 (2010).
[16] Gazzah, S., Mhalla, A. and Amara, N. E. B., "Vehicle detection on a video traffic scene: Review and new perspectives," Information and Digital Technologies (IDT), 2017 International Conference on. IEEE, 448-454(2017).
[17] He, K., Zhang, X., Ren, S. and Sun, J., "Deep residual learning for image recognition," Proc. CVPR, 770-778(2016).
[18] Qiu, Z., Yao, T., & Mei, T., "Learning spatio-temporal representation with pseudo-3d residual networks," Proc. ICCV, 5534-5542 (2017).
[19] Deng, J., Dong, W., Socher, R. and Li, L. J., "Imagenet: A large-scale hierarchical image database," Proc. CVPR, 248-255 (2009).
[20] Simonyan, K., Zisserman, A., "Very deep convolutional networks for large-scale image recognition," arXiv preprint arXiv: 1409.1556 (2014).